\documentclass[letterpaper, 10 pt, conference]{ieeeconf}

\IEEEoverridecommandlockouts
\usepackage{booktabs}
\usepackage{multirow}
\usepackage{graphicx}
\usepackage{amsmath}
\usepackage{graphicx}
\usepackage{subcaption}
\usepackage{amssymb}
\usepackage{algorithm}
\usepackage{caption}
\usepackage{mathrsfs}
\usepackage{amsfonts}
\usepackage{algpseudocode}
\usepackage{mathtools}
\usepackage{mathalfa}

\usepackage[most]{tcolorbox}

\begin{document}

\title{\LARGE \bf
CAMO: A Conditional Neural Solver for the Multi-objective Multiple Traveling Salesman Problem
}

\author{
Fengxiaoxiao Li$^{1,*}$, Xiao Mao$^{2,*}$, Mingfeng Fan$^{1,\dag}$, Yifeng Zhang$^{1}$, Yi Li$^{3}$, 
 \\
Tanishq Duhan$^{1}$, and Guillaume Sartoretti$^{1}$
\thanks{\textsuperscript{*}Equal contribution. \textsuperscript{\dag}Corresponding author: ming.fan@nus.edu.sg}
\thanks{$^{1}$National University of Singapore, Singapore.}%
\thanks{$^{2}$China Unicom Shenzhen Branch, Shenzhen, China.}%
\thanks{$^{3}$Central South University, Changsha, China.}%
}

\maketitle
\thispagestyle{empty}
\pagestyle{empty}
\begin{abstract}

Robotic systems often require a team of robots to collectively visit multiple targets while optimizing competing objectives, such as total travel cost and makespan. This setting can be formulated as the Multi-Objective Multiple Traveling Salesman Problem (MOMTSP). Although learning-based methods have shown strong performance on the single-agent TSP and multi-objective TSP variants, they rarely address the combined challenges of multi-agent coordination and multi-objective trade-offs, which introduce dual sources of complexity.
To bridge this gap, we propose CAMO, a conditional neural solver for MOMTSP that generalizes across varying numbers of targets, agents, and preference vectors, and yields high-quality approximations to the Pareto front (PF).
Specifically, CAMO consists of a conditional encoder to fuse preferences into instance representations, enabling explicit control over multi-objective trade-offs, and a collaborative decoder that coordinates all agents by alternating agent selection and node selection to construct multi-agent tours autoregressively.
To further improve generalization, we train CAMO with a REINFORCE-based objective over a mixed distribution of problem sizes. Extensive experiments show that CAMO outperforms both neural and conventional heuristics, achieving a closer approximation of PFs. In addition, ablation results validate the contributions of CAMO's key components, and real-world tests on a mobile robot platform demonstrate its practical applicability.




\end{abstract}

\section{Introduction}

The Multiple Traveling Salesman Problem (MTSP) aims to plan routes for multiple agents such that each target location is visited exactly once, while optimizing a predefined objective over the resulting tours. MTSP arises naturally in multi-robot systems that require coordinated task allocation and route planning under practical constraints. Representative applications include cooperative dual-arm rearrangement~\cite{li2024synchronized}, constrained multiple-depot routing~\cite{yang2024hierarchical}, multi-platform agricultural operations~\cite{carpio2021mp}, and cooperative multi-agent navigation in dynamic environments~\cite{liang2024hd}.
Based on the objective definition, MTSP is commonly categorized into Min-Sum MTSP, which minimizes the total tour length, and Min-Max MTSP, which minimizes the longest tour length (i.e., the makespan). Both variants are NP-hard to solve optimally~\cite{guo2024imtsp} and belong to the class of single-objective combinatorial optimization problems (SOCOPs).

In practice, however, decision-makers often need to balance competing goals; for example, minimizing total travel distance may worsen workload imbalance among robots. Such requirements motivate the Multi-Objective MTSP (MOMTSP)~\cite{bektas2006multiple}, a representative multi-agent multi-objective combinatorial optimization problem (MOCOP), which seeks a diverse set of Pareto-optimal solutions under different preferences that capture trade-offs among conflicting objectives. MOMTSP therefore poses dual challenges: coordinating multiple agents while simultaneously modeling diverse trade-offs among objectives.

Traditional Multi-Objective Evolutionary Algorithms (MOEAs)~\cite{zhou2011multiobjective} have been widely adopted to approximate the Pareto front (PF) to handle MOMTSP.
However, their population-based iterative search often incurs high computational cost and limited scalability when handling large-scale problem instances. 
Moreover, many approaches rely on handcrafted operators or problem-specific heuristics, which require substantial expert knowledge and reduce adaptability across different robotic scenarios~\cite{coello2007evolutionary}.
As robotic systems increasingly require real-time responsiveness to dynamic events (e.g., newly appearing targets or robot failures), existing approaches~\cite{shim2012hybrid} struggle to meet these demands simultaneously.

In recent years, learning-based methods, particularly deep reinforcement learning (DRL), have achieved strong performance on combinatorial optimization problems such as single-agent TSP~\cite{guo2024imtsp} and multi-objective TSP. However, extending these neural solvers to MOMTSP is challenging because it requires handling both multi-objective trade-offs and multi-agent coordination within a single policy. On the multi-objective side, the policy must adapt to different preference vectors and produce solutions that cover diverse regions of the Pareto front, rather than optimizing a single scalar reward. 
On the multi-agent side, the model must coordinate a variable number of agents under a \emph{predefined team size} and handle \emph{agent idling} consistently (e.g., some agents may remain at the depot under preferences that prioritize minimizing total tour length), while still producing an executable multi-agent plan with well-defined routes for all agents. This requirement is not captured by directly extending single-agent formulations or by models that assume flexible vehicle usage, as in multi-objective capacitated vehicle routing problems (MOCVRP)~\cite{fan2025preference}.

To address these limitations, we propose CAMO (\underline{C}onditional \underline{A}ttention for \underline{M}ulti-objective \underline{O}ptimization), a conditional neural solver for MOMTSP. 
CAMO adopts a specialized encoder–decoder architecture: a conditional encoder fuses preference and instance features to model multi-objective trade-offs, while a collaborative decoder alternates between an agent-selection module and a node-selection module to construct tours. The decoder respects a predefined team size and scales naturally to varying numbers of agents. In particular, the agent-selection module coordinates the team by attending to a global context and all agents’ current states when selecting which agent acts next, thereby enabling effective multi-agent collaboration.
The main contributions are summarized as follows:
\begin{itemize}
    \item We propose CAMO, a conditional neural solver for MOMTSP that jointly addresses multi-objective trade-offs and multi-agent coordination, and generalizes across varying numbers of targets, agents, and preference vectors. We further train CAMO with a REINFORCE-based objective over a mixed distribution of problem sizes to improve cross-scale generalization.
    \item We design a conditional encoder with conditional attention (CA) to integrate preference and instance information, together with a gated aggregation (GA) module to enhance representations, and develop a collaborative decoder that coordinates agents via global–agent state attention with alternating agent and node selection.
    \item We conduct extensive experiments to demonstrate CAMO’s effectiveness and validate real-world applicability through physical tests on a mobile robot platform.
\end{itemize}

\section{Related Works}

\noindent \textbf{Traditional methods for MOCOPs.}
MOEAs are the cornerstone metaheuristic for MOCOPs~\cite{deb2011multi}. They can be broadly categorized into dominance-based methods, such as NSGA-II~\cite{deb2002fast}, 
 which combine 
Pareto ranking with crowding-distance preservation. Decomposition-based approaches, such as MOEA/D~\cite{zhang2007moea}, which scalarize objectives 
into collaborative single-objective subproblems. Subsequent advances include indicator-based selection like IBEA~\cite{zitzler2004indicator}, which guides search via hypervolume or $\epsilon$-dominance, and reference-point-based methods 
like NSGA-III~\cite{deb2013evolutionary}, designed to maintain uniform front 
coverage in many-objective settings where dominance pressure deteriorates. 
Problem-specific evolutionary approaches have also been explored for MOMTSP, such as hybrid EDAs~\cite{shim2012hybrid} that exploit a probabilistic model 
building over permutation spaces.

\noindent \textbf{Neural Multi-Objective Combinatorial Optimization.}
Existing neural solvers for MOCOPs primarily adopt a decomposition-based 
strategy~\cite{zhang2007moea}, scalarizing the multi-objective problem into 
weighted single-objective subproblems parameterized by a preference vector 
$\lambda$. While early works trained a distinct model for each preference, 
suffering from inflexibility and high training costs, recent SOTA methods favor 
a single, generalized model conditioned on $\lambda$, enabling continuous 
preference interpolation at inference time. The central challenge is effectively 
integrating this preference signal; approaches range from hypernetworks like 
PMOCO~\cite{lin2022pareto}, which generate task-specific weights from $\lambda$, 
to encoder-fusion methods like CNH~\cite{fan2024conditional} and 
WE-CA~\cite{chen2025rethinking}, which inject preference information directly 
into the attention mechanism. However, these representative works, including 
prominent collaborative DRL methods~\cite{wu2024collaborative,fan2025preference}, 
are exclusively designed for single-agent MOCOPs and fundamentally cannot address 
the cooperation  required in 
multi-agent settings.

\section{Background}

\subsection{Problem Formulation}
The MOMTSP problem is formally defined on a complete graph $G(\Xi, E)$. The node set $\Xi = \{0\} \cup \Xi'$ consists of a single depot indexed by $0$ (with location $(x_0, y_0)$) and a set of $n_\xi$ nodes $\Xi' = \{(x_1, y_1), \dots, (x_{n_\xi}, y_{n_\xi})\}$. A fleet of $n_a$ agents $\mathcal{A} = \{1, \dots, n_a\}$ is available at the depot. 
The goal is to determine a set of routes, one for each agent, such that every agent starts and ends at the depot, and every node in $\Xi'$ is visited exactly once by exactly one agent.    
In our study, we consider two objectives: $f_1=\min \sum_{a\in\mathcal{A}}\sum_{i\in\Xi}\sum_{j\in\Xi} w_{ij}\psi_{ij}^{a}$, which minimizes the total tour length of all agents; and $f_2=\min \max_{a\in\mathcal{A}} r_a$, which minimizes the makespan. Here, $w_{ij}$ denotes the distance between nodes $i$ and $j$, $\psi_{ij}^{a}\in\{1,0\}$ indicates whether agent $a$ travels from $i$ to $j$, and $r_a$ represents the cumulative travel distance of agent $a$.

MOMTSP is challenging due to its NP-hardness, involving two coupled 
decisions: assigning nodes to agents and determining each agent's visiting order. 
For instance, an instance with $100$ cities and $4$ agents already yields 
 $100! \times \binom{99}{3} = 
\frac{100!\times 99!}{3!\times 96!}\approx 10^{163}$ candidate solutions. 
Beyond this, introducing two objectives fundamentally transforms the search: 
rather than finding a single optimum, one must identify the PF
of non-dominated solutions. Verifying Pareto optimality requires pairwise 
comparisons across all candidates---up to $O\!\left(10^{326}\right)$ in the 
worst case---rendering exhaustive search utterly infeasible, dwarfing even the 
estimated number of atoms in the observable universe ($10^{78}$--$10^{82}$).

\textbf{Definition 1 (Pareto Dominance).}
For a problem with $n$ objectives, let $\Pi$ comprise the set of all feasible solutions. 
A solution $\pi \in \Pi$ is said to dominate another solution $\pi' \in \Pi$ (denoted $\pi \prec \pi'$)  only if:
$
\forall i \in \{1, \dots, n\}, f_i(\pi) \le f_i(\pi')$ and $ \exists j \in \{1, \dots, n\}, f_j(\pi) < f_j(\pi')
$.

\textbf{Definition 2 (Pareto Optimality).}
A solution $\pi^* \in \Pi$ is Pareto optimal if it is not dominated by any other solution $\pi \in \Pi$. 
The set of all solutions is the \emph{Pareto set}, $\mathcal{P} = \{\pi^* \in \Pi \mid \nexists \pi \in \Pi, \pi \prec \pi^*\}$. The image of this set in the objective space is the  PF, $\mathcal{F} = \{F(\pi^*) \mid \pi^* \in \mathcal{P}\}$, where 
 $F(\pi) = (f_1(\pi), \ldots, f_n(\pi))^\top$ maps each solution to a point.


\subsection{Objective Decomposition Strategy}

Decomposition strategies are widely used to solve MOCOPs~\cite{zhou2011multiobjective}, which reformulate the MOCOP, $F(\pi) = (f_1(\pi), \dots, f_n(\pi))^\top$, as a set of single-objective combinatorial optimization problems (SOCOPs).
Each SOCOP is parameterized by a preference (weight) vector $\lambda = (\lambda_1, \dots, \lambda_n)$, where $\lambda_j \ge 0$.
By solving these SOCOPs with diverse $\lambda$ vectors, an approximation of the PF is obtained.

Given the ability to find solutions on non-convex PFs, we adopt the Tchebycheff (TC) decomposition to scalarize the MOMTSP~\cite{lin2022pareto}.
This function minimizes the maximum weighted deviation from an ideal objective vector $z^* = (z_1^*, \dots, z_n^*)$. 
For a given preference vector $\lambda_i$, a Single Objective MTSP (SOMTSP) is defined as:
\begin{equation}
    \min_{\pi} g^{tc}(\pi | \lambda_i, z^*) = \max_{j=1,\dots,n} \{ \lambda_i^j |f_j(\pi) - z_j^*| \}.
\end{equation}

\section{CAMO}

\subsection{Overview} 
\begin{figure*}[h]
    \centering
    \includegraphics[width=\linewidth]{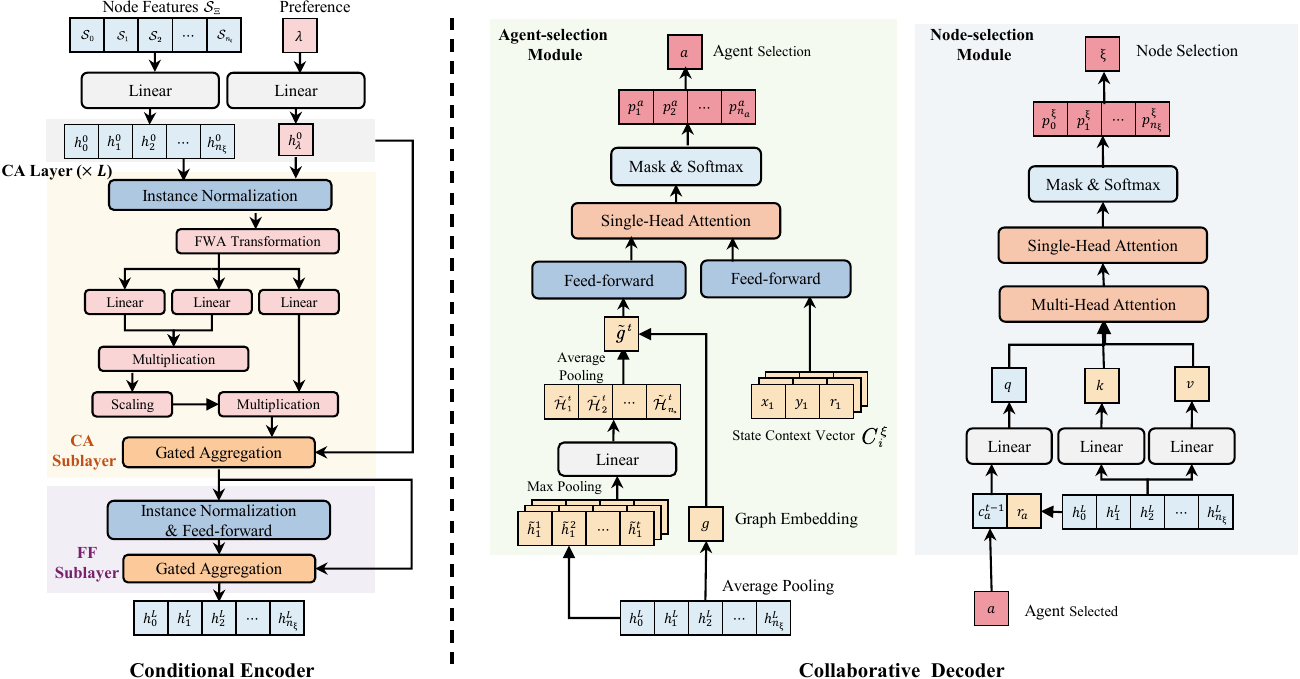}
    \caption{The network architecture of CAMO.}
    \label{fig:network architecture of CAMO}
\end{figure*}

To achieve high-quality approximate Pareto optimal solutions for MOMTSP, we propose a conditional neural solver, CAMO, conditioned on the instance context $\mathcal{I}$, preference $\lambda$, agent counts $n_a$, and node scales $n_\xi$.
Our CAMO is designed to learn a stochastic policy $p_\theta$ for obtaining the approximate output solution. 
The final output is a sequence of action tuples $(a_t, \xi_t)$, consisting of selected agents and nodes. 
The output solution is denoted as $\pi = \{(a_0, \xi_0), \dots, (a_T, \xi_T)\}, \ \xi_t \in \Xi, \ a_t \in \mathcal{A}$, expressed as:
\begin{equation}
\begin{split}
    p_\theta(\pi \mid \mathcal{I}, \lambda, n_a, n_\xi) &= \prod_{t=0}^{T} p_\theta \Big( (a_t, \xi_t) \mid \mathcal{I}, \lambda, n_a, n_\xi, \\
    &\qquad \qquad \qquad (a_{0:t-1}, \xi_{0:t-1}) \Big),
\end{split}
\end{equation}
where $\theta$ represents the set of learnable parameters, and $(a_t, \xi_t)$ and $(a_{0:t-1}, \xi_{0:t-1})$ represent the selected agent--node action tuple and the partial solution at time step $t$, and $T$ denotes the number of steps to construct the solution $\pi$.
To this end, we utilize an encoder-decoder architecture. 
The conditional encoder embeds and fuses the instance context~$\mathcal{I}$ and preference~$\lambda$, allowing the model to understand their combined impact on the SOMTSPs and the eventual PF.
The collaborative decoder, comprising iterative agent-selection and node-selection modules, provides dual scalability and generalizes to unseen agent counts $n_a$ and node scales $n_\xi$, offering an advantage over inflexible classifier-based structures~\cite{xu2021reinforcement}. Moreover, the agent-selection module coordinates multi-agent decision making via global–agent state attention, enabling effective collaboration.
The neural architecture of our CAMO is illustrated in Fig.~\ref{fig:network architecture of CAMO}. 
We elaborate on the details of these key components in the following sections.

\subsection{Conditional Encoder}

The conditional encoder takes an instance $\mathcal{I}$ and a preference $\lambda$ as inputs and outputs enhanced node embeddings, generated through a stack of $L$ CA layers. 
Each layer consists of a CA sub-layer and a feed-forward (FF) sub-layer.

\noindent\textbf{Raw Features.}
The conditional encoder first projects raw node features $\mathcal{S}_\Xi = \{\mathcal{S}_0, \mathcal{S}_1,..., \mathcal{S}_{n_\xi}\}$ and the preference $\lambda$ into initial node embeddings $H_\xi^0 = \{h_0^0, \dots,  h_{n_\xi}^0\}$ and a preference embedding $h_\lambda^0$. 
Passing these directly to an Attention Model (AM)~\cite{kool2018attention} encoder would indiscriminately blend heterogeneous inputs and obscure their specific interdependencies. To address this, we introduce a CA mechanism that conditions node embeddings on the preference vector and jointly updates preference and node embeddings to better capture preference--node interactions, thereby improving the solution quality. At layer $l \in \{1,\cdots,L\}$, the CA layer receives the node embeddings $H_\xi^{\,(l-1)} = \{h_0^{\,(l-1)}, \dots, h_{n_\xi}^{\,(l-1)}\}$ and a preference embedding $h_{\lambda}^{\,(l-1)}$ as inputs.

\noindent\textbf{CA Sublayer.}
As illustrated in Fig. \ref{fig:network architecture of CAMO}, the node embeddings 
$H_\xi^{\,(l-1)} = \{h_0^{\,(l-1)},  \dots, h_{n_\xi}^{\,(l-1)}\}$ 
and the preference embedding $h_{\lambda}^{\,(l-1)} $ are first passed through Instance Normalization (IN)~\cite{ulyanov2016instance}: $h_{i}^{*{(l-1)}} = \text{IN} (h_{i}^{{(l-1)}}), h_{\lambda}^{*{(l-1)}} = \text{IN} (h_{\lambda}^{{(l-1)}})$. Then, we condition the normalized node embeddings $h_{i}^{*{(l-1)}}$ on the normalized preference embedding $h_{\lambda}^{*(l-1)}$ via a feature-wise affine (FWA) transformation~\cite{perez2018film}, as follows:
\begin{equation}
\begin{gathered}
    \gamma^{(l-1)} = W^{\gamma}h_{\lambda}^{*(l-1)}, \quad \beta^{(l-1)} = W^{\beta}h_{\lambda}^{*(l-1)}, \\
    h_{i}^{\prime{(l-1)}} = \gamma^{(l-1)} \circ h_{i}^{*(l-1)} + \beta^{(l-1)}, \quad \forall i \in \{0,...,n_\xi\},
\end{gathered}
\end{equation}
where $W^{\gamma}$ and $W^{\beta}$ are trainable matrices, and $\circ$ denotes element-wise multiplication. The resulting node embeddings $h_{i}^{\prime{(l-1)}}$ and the normalized preference embedding $h_{\lambda}^{*(l-1)}$ are then jointly updated with a standard multi-head attention (MHA) block to encourage information exchange and mitigate representational conflicts. Let $\mathcal{C}^{(l-1)}=\{h_{\lambda}^{*(l-1)}\}\cup\{h_{j}^{\prime(l-1)}\}_{j=0}^{n_\xi}$ denote the context used as Key/Value vectors:
\begin{align}
    \hat{h}_{\lambda}^{(l-1)} &= \text{MHA}(q=h_{\lambda}^{*(l-1)}, k=\mathcal{C}^{(l-1)}, v=\mathcal{C}^{(l-1)}), \\
    \begin{split}
        \hat{h}_{i}^{(l-1)} &= \text{MHA}(q=h_{i}^{\prime(l-1)}, k=\mathcal{C}^{(l-1)}, v=\mathcal{C}^{(l-1)}), \\
        &\qquad \forall i \in \{0,...,n_\xi\},
    \end{split}
\end{align}
where $q$, $k$, and $v$ represent the Query, Key and Value vectors in MHA, respectively. Finally, we replace standard additive residual connections~\cite{szegedy2017inception} with GA~\cite{li2024moganet} to regulate information flow and enhance expressive power~\cite{srivastava2015training}. We apply GA at the end of the CA sublayer:
\begin{equation}
\operatorname{GA}(\mathcal{X}, \mathcal{Y})
= \mathcal{X} + \sigma ( U_g\,\mathcal{X} - b_g )\odot \mathcal{Y},
\end{equation}
where $\mathcal{X} = \{h_{\lambda}^{(l-1)}\} \cup \{h_j^{(l-1)}\}_{j=0}^{n_\xi}$ denotes the inputs to the CA sublayer, and $\mathcal{Y} = \{\hat{h}_{\lambda}^{(l-1)}\} \cup \{\hat{h}_j^{(l-1)}\}_{j=0}^{n_\xi}$ denotes the outputs of the MHA block.
Besides, $U_g$ and $b_{g}$ are trainable parameters, $\sigma$ is the Sigmoid function~\cite{hopfield1984neurons}, and $\odot$ denotes element-wise multiplication.

\noindent\textbf{FF Sublayer.} 
The FF sublayer takes the outputs of the CA sublayer as input. It comprises Instance Normalization (IN), a two-layer MLP~\cite{shazeer2020glu} with ReLU activation~\cite{nair2010rectified}, and a dedicated GA module, applied sequentially. For the GA in the FF sublayer, we set $\mathcal{X}$ to the inputs of the FF sublayer and $\mathcal{Y}$ to the MLP outputs.
The final node embeddings $H_\xi^L = \{h_0^L, \dots, h_{n_\xi}^L\}$  produced by the conditional encoder serve as inputs to the collaborative decoder.

\subsection{Collaborative Decoder}
The collaborative decoder, composed of an agent-selection module and a node-selection module, takes as input the final node embeddings produced by the conditional encoder, together with a graph embedding $g$ obtained by average pooling over these node embeddings. It constructs the solution auto-regressively. At each decoding step, the decoder sequentially utilizes the agent-selection module and then the node-selection module, iteratively generating a sequence of agent-node tuples to build the complete multi-agent solution.

\noindent\textbf{Agent-selection Module.}
At decoding step $t$, we construct the current node sequence embeddings for all agents. Let $\tilde{H}_i^t = \{\tilde{h}_i^0, \dots, \tilde{h}_i^{t-1}\}$ denote the sequence of  node embeddings visited by agent $i$ before step $t$. The set of all sequences is $\mathcal{H}^t = \{\tilde{H}_1^t, \dots, \tilde{H}_{n_a}^t\}$.

This set $\mathcal{H}^t$ is processed via max pooling and a linear layer to obtain an aggregated embedding:
$\tilde{\mathcal{H}}^t = \text{maxpool}(\mathcal{H}^t) W_a + b_a$,
where $W_a$ and $b_a$ are trainable parameters. Here,
$\tilde{\mathcal{H}}^t = {\{\tilde{\mathcal{H}}_1^t, \dots, \tilde{\mathcal{H}}_{n_a}^t\}}$
denotes the set of current node-sequence embeddings for all $n_a$ agents at step $t$. Subsequently, we compute a global representation $\tilde{g}^{t}$ by performing average pooling over these per-agent embeddings and then fusing the result with the graph embedding $g$, that is,
$\tilde{g}^t = \frac{1}{n_a} \sum_{i=1}^{n_a} \tilde{\mathcal{H}}_i^t + g$.

Concurrently, we define a state context vector $C_i^\xi = (x_i, y_i; r_i)$ for each agent $i$, which captures its current coordinates $(x_i, y_i)$ and cumulative travel distance $r_i$. The global representation $\tilde{g}^t$ and each agent's state context $C_i^\xi$ are then processed through separate FF layers to generate the final global Query vector $\mathcal{G}^t$ and the agent-specific Key vectors $\hat{\mathcal{G}_i^t}$, respectively:
\begin{equation}
\mathcal{G}^t = \text{FF}(\tilde{g}^t); \quad \hat{\mathcal{G}_i^t} = \text{FF}(C_i^\xi) \quad \forall i \in \{1, 2, \dots, n_a\}.
\end{equation}
These FF layers share the same architecture as those in the encoder but utilize independently trained parameters.

Finally, we use $\mathcal{G}^t$ as the Query and $\hat{\mathcal{G}_i^t}$ as the Key to compute the attention score $s_i^a$ using a single-head attention (SHA) layer:
\begin{equation}
s_i^a = \kappa \cdot \tanh((\hat{\mathcal{G}_i^t})^\top \mathcal{G}^t / \sqrt{d_k}) \quad \forall i \in \{1, 2, \dots, n_a\},
\end{equation}
where $\kappa$ is a clipping coefficient to ensure numerical stability and $d_k$ is the dimensionality of the Key vector. After calculating the scores $s_i^a$, agents that have already returned to the depot are masked. A softmax function is then applied to the attention scores to yield the agent selection probabilities $\{p_1^{a},\, p_2^{a},\, \ldots,\, p_{n_a}^{a}\}$, thereby determining the agent $a_t$ selected at step $t$. This design enables the module to adapt to varying scales of agents, ensuring model scalability.

\noindent\textbf{Node-selection Module.}
Once the agent $a_t$ is selected, the node-selection process begins. We construct a node context vector $C_a^t = [c_a^{t-1}; r_a]$, where $c_a^{t-1}$ is the embedding of the node visited by agent $a_t$ at the previous step $t-1$, and $r_a$ is its current travel distance. The selection is determined using an MHA block followed by a SHA layer.

In the MHA  with $M$ heads, the Query ($q_m^\xi$) is derived from the node context $C_a^t$, while the Key ($k_{i,m}^\xi$) and Value ($v_{i,m}^\xi$) are derived from the final node embeddings ${h}_i^L$:
\begin{equation}
\begin{split}
q_m^\xi &= C_a^t W_{q,m}^\xi; \quad k_{i,m}^\xi = {h}_i^L W_{k,m}^\xi; \\
v_{i,m}^\xi &= {h}_i^L W_{v,m}^\xi \quad \forall i \in \{0, \dots, n_\xi\},
\end{split}
\end{equation}
where $W_{q,m}^\xi$, $W_{k,m}^\xi$, and $W_{v,m}^\xi$ are the trainable parameter matrices for head $m \in \{1, 2, \cdots, M\}$. Next, the attention score $e_{i,m}^\xi$ is calculated, and a mask is applied to visited nodes:
\begin{equation}
e_{i,m}^\xi = q_m^\xi (k_{i,m}^\xi)^{\top} / \sqrt{d_k} \quad \forall i \in \{0, \dots, n_\xi\},
\end{equation}
\begin{equation}
e_{i,m}^{\xi} = \begin{cases} e_{i,m}^\xi, & \text{if node } i \text{ is unvisited,} \\ -\infty, & \text{if node } i \text{ is visited.} \end{cases}
\end{equation}
The scores $e_{i,m}^\xi$ are normalized via softmax to obtain attention weights $\alpha_{i,m}^\xi$. These weights are used to compute the head output $z_m^\xi$. The outputs of all heads are concatenated and passed through a linear mapping $W^\xi$ to yield the advanced node context $q^\xi$:
\begin{equation}
\begin{split}
\alpha_{i,m}^\xi &= \text{softmax}(e_{i,m}^\xi); \quad
z_m^\xi = \sum_{i=0}^{n_\xi} \alpha_{i,m}^\xi v_{i,m}^\xi; \\
q^\xi &= [z_1^\xi, z_2^\xi, \dots, z_M^\xi]W^\xi,
\end{split}
\end{equation}
where $W^\xi \in \mathbb{R}^{d_h \times d_h}$ is a trainable parameter matrix.

The advanced node context $q^\xi$ is then used as the Query, while the final node embeddings serve as the Key vector in the subsequent SHA layer. The attention score $s_i^\xi$ in SHA is calculated as:
\begin{equation}
s_i^\xi = \kappa \cdot \tanh(q^\xi ({h}_i^L)^\top / \sqrt{d_k}) \quad \forall i \in \{0, 1, \dots, n_\xi\}.
\end{equation}
A masking mechanism is applied to the scores $s_i^\xi$, and a softmax function is used to obtain the final node selection probabilities $\{p_0^\xi, p_1^\xi, \dots, p_{n_\xi}^\xi\}$ for agent $a_t$. 
The final node $\xi_t$ is chosen based on these probabilities using either a greedy or sampling strategy.
During training, we adopt a customized sampling strategy that applies to both agent and node selection, thereby encouraging exploration. During inference, a greedy strategy is employed to maximize the expected reward. 
Crucially, the parameters in the agent and node selection decoders are independent of the number of agents and nodes.
This ensures the policy network is scalable and can be applied to problems of varying scales.

\subsection{Training Algorithm}

We train the policy network $p_\theta$ using a REINFORCE-based objective to obtain approximate Pareto solutions for MOMTSP. The training procedure is augmented with a customized sampling strategy that, for each batch, samples different agent counts $n_a$, node scales $n_\xi$, and preference vectors $\lambda$. Given an MOMTSP instance $\mathcal{I}$ with preference $\lambda$, we aim to maximize the expected return $J(\theta) = \mathbb{E}_{\pi \sim p_\theta(\pi \mid \mathcal{I}, \lambda, n_a, n_\xi)} [R(\pi)]$, where the reward $R(\pi) = g^{tc}(\pi \mid \lambda_i, z^*)$ is derived from the TC decomposition. To reduce the variance of the policy gradient, we adopt a baseline $R_b$:
\begin{equation}
\begin{split}
    \nabla_\theta J(\theta) &= \mathbb{E}_{\pi \sim p_\theta(\pi \mid \mathcal{I}, \lambda, n_a, n_\xi)} \Big[ (R(\pi) - R_b) \\
    &\quad \times \nabla_\theta \log p_\theta(\pi \mid \mathcal{I}, \lambda, n_a, n_\xi) \Big].
\end{split}
\raisetag{15pt} 
\end{equation}
For a batch of $B$ instances, we sample $K$ sequences $\pi_b^k$ for each instance $\mathcal{I}_b$ using the POMO strategy~\cite{kwon2020pomo}, and calculate the approximate gradient as below:
\begin{equation}
\label{eq:Reinforce}
\begin{split}
    \nabla_\theta {J(\theta)} &\approx \frac{1}{BK} \sum_{b=1}^B \sum_{k=1}^K \Big[ (R(\pi_b^k) - R_b) \\
    &\quad \times \nabla_\theta \log p_\theta(\pi_b^k \mid \mathcal{I}_b, \lambda_b, n_a^b, n_\xi^b) \Big],
\end{split}
\end{equation}
where the baseline $R_b = \frac{1}{K} \sum_{k=1}^K R(\pi_b^k)$ is the average reward of $K$ sequences sampled for instance $\mathcal{I}_b$, where each sequence starts with a different first node. 
The full training process is detailed in Algorithm~\ref{alg:reinforce}.

\begin{algorithm}[h]
    \caption{\textbf{Training Algorithm}}
\label{alg:reinforce}
\small
\begin{algorithmic}[1]
\Require Number of training epochs $\mathbb{E}$, training rounds $T$, batch size $B$, sampling sequences $K$, agent counts set $\mathcal{D}_\mathcal{A}$, node scales set $\mathcal{D}_{\Xi}$
\Ensure Policy network $p_\theta$
\State Initialize policy network parameters $\theta$
\For{$e = 1$ to $\mathbb{E}$}
    \For{$t = 1$ to $T$}
        \State $\mathcal{I}_b \gets \text{ProblemSampler}()$, 
        $\forall b \in \{1,\dots,B\}$
        \State $n_a \gets \text{AgentCountSampler}(\mathcal{D}_\mathcal{A})$
        \State $n_\xi \gets \text{NodeScaleSampler}(\mathcal{D}_{\Xi})$
        \State $\lambda \gets \text{PreferenceSampler}()$
        \State $\pi_b^k \gets \text{TourSampler}\bigl(p_\theta(\cdot \mid \mathcal{I}_b,\lambda,n_a,n_\xi)\bigr)$,
        \Statex \hspace{4.5em} $\forall b \in \{1,\dots,B\},\ \forall k \in \{1,\dots,K\}$
        \State $R_b \gets \frac{1}{K}\sum_{k=1}^{K} R(\pi_b^k)$, 
        $\forall b \in \{1,\dots,B\}$
        \State $\nabla_\theta J(\theta) \gets \frac{1}{BK}
        \sum_{b=1}^{B} \sum_{k=1}^{K}
        \bigl[(R(\pi_b^k)-R_b)$
        \Statex \hspace{6em} $\cdot \nabla_\theta \log p_\theta(\pi_b^k \mid \cdot)\bigl]$
        \State $\theta \gets \text{Adam}(\theta, \nabla_\theta J(\theta))$
    \EndFor
\EndFor
\end{algorithmic}
\end{algorithm}

\section{Experiments}

\subsection{Experimental settings}

\noindent\textbf{Problems and Instance Generation.}
We evaluate CAMO on MOMTSP instances where all node and depot locations are randomly generated within a $[0, 1]^2$ unit square. 
We evaluate CAMO under different node scales and agent counts, including 20 nodes with 2 agents, 50 nodes with 3 agents, and 100 nodes with 4 agents, denoted as N20A2, N50A3, and N100A4, respectively. 

To boost generalization across varying problem sizes, we employ a customized sampling strategy. 
Specifically, for each training batch, we randomly sample the preference vector $\lambda$ by drawing each element from $[0,1]$ to induce diverse stochastic preferences~\cite{fan2024conditional}. When computing rewards, we normalize $\lambda$ to sum to 1, so that it represents the relative importance of the objectives. 
The node scale $n_\xi$ and agent number $n_a$ are randomly drawn from the integer ranges $[20, 100]$ and $[2, 4]$, respectively. 

\noindent\textbf{Hyperparameters.}
We train CAMO for $\mathbb{E}=400$ epochs, sampling $N=100,000$ instances per epoch. 
We use the Adam optimizer with a batch size $B=64$, a learning rate of $10^{-4}$, and a weight decay of $10^{-6}$.
The solution sampling size per instance $K$ is set to the node scale, $K=n_\xi$.
For the network architecture, the embedding dimension is $d_h=128$. The MHA employs $M=8$ heads with Query and Key dimensions of $16$. 
The hidden dimension of the FF sublayer is $d_f=512$. The encoder consists of $L=6$  CA layers, and the clipping coefficient $\kappa$ is 10.


\noindent\textbf{Baselines.}
We compare CAMO against a range of established baseline methods, including both classical MOEAs and a representative neural solver.
Specifically, the selected traditional MOEAs include:
1) \textbf{MOEA/D}~\cite{zhang2007moea}, a classical decomposition-based MOEA, where the neighborhood size, neighborhood selection probability, and the maximum number of iterations are set to 15, 0.7, and 4{,}000, respectively;
2) \textbf{MOGLS}~\cite{chen2014new}, a multi-objective genetic local search algorithm with a population size of 140, executed for 10{,}000 iterations with 30 local search steps per iteration;
3) \textbf{NSGA-II}~\cite{deb2002fast}, a widely used Pareto dominance-based genetic algorithm with a population size of 100 and 4{,}000 iterations; and
 4) \textbf{NSGA-III}~\cite{deb2013evolutionary}, an extension of NSGA-II incorporating reference directions, also run for 4{,}000 iterations.
Inspired by the work in~\cite{lacomme2006genetic}, we apply problem-specific local operators for solving MOMTSP.
For the DRL-based category, we benchmark against 5) \textbf{MO-PARCO}: we adapt PARCO~\cite{berto2024parco}, a SOTA neural solver originally designed for multi-agent SOCOPs, where agents construct solutions in parallel through coordinated decision-making. To make it applicable to the multi-objective setting, we extend PARCO by conditioning the policy on preference vectors via the same CA mechanism.


\noindent\textbf{Inference and Evaluation Metrics.} During inference, we employ an Instance Augmentation (IA) strategy~\cite{kwon2020pomo} for CAMO (i.e, CAMO-IA), solving 8 transformations of each instance and reporting the solution with the best HV as the final result. 
We evaluate all methods using three metrics: Hypervolume (HV)~\cite{zitzler2003performance}, Gap, and total runtime (Time).
HV is a widely used indicator for measuring the coverage of the PF, where a larger HV value indicates better performance.
Given a set $P \subset \mathbb{R}^n$ and a reference point $\mathfrak{r}^{*}$ dominated by all solutions in $P$, the HV is defined as the Lebesgue measure of the region it dominates up to the reference point $\mathfrak{r}^{*}$: $HV(P)=\zeta^{n}\!\left(\left\{\mathfrak{r}\in\mathbb{R}^{n}\,\middle|\,\exists\,u\in P \text{ such that } u \prec \mathfrak{r} \prec \mathfrak{r}^{*}\right\}\right)$, where $\zeta^{n}$ denotes $n$-dimensional Lebesgue measure (i.e., the volume in $\mathbb{R}^{n}$). We report normalized HV values in $[0,1]$ with the same $\mathfrak{r}^{*}$ across all methods.
In our experiments, for each problem scale, we set the reference point to $1.1$ times the maximum objective value observed across all methods, ensuring that it is dominated by the union of their solution sets. For convenience, we then round the reference point up to the nearest integer.
The Gap metric quantifies the relative performance difference between a given method and CAMO-IA in terms of HV, formally defined as $\mathrm{Gap} = (HV_{\text{CAMO-IA}} - HV_i)/HV_{\text{CAMO-IA}} \times 100\%$, where $HV_i$ denotes the HV of the $i$-th method. A smaller Gap indicates closer performance to CAMO-IA.

All experiments are implemented in Python and conducted on a device equipped with an NVIDIA Ampere A100 GPU and an AMD EPYC 7742 CPU.
To ensure reproducibility, our code implementation and datasets will be made publicly available upon paper acceptance.





\subsection{Experimental results}

\noindent\textbf{Comparison Analysis.}
We evaluate all algorithms on 100 randomly generated instances for each configuration, including N20A2, N50A3, and N100A4. 
Since MOMTSP lacks standard a priori reference points for HV calculation, we establish them via conservative estimation from the complete solution pool of all methods.
The reference points are set to [7, 6] for N20A2, [14, 11] for N50A3, and [29, 20] for N100A4, respectively. 
This setup ensures that all obtained non-dominated solutions are covered, preventing evaluation bias in the HV metric.

\begin{table}[ht]
    \centering
    \caption{Baseline comparison results.}
    \label{tab:Comparison Results}
    
    \renewcommand{\arraystretch}{1.0} 
    \setlength{\tabcolsep}{1.5pt}     
    
    \resizebox{\columnwidth}{!}{
        \begin{tabular}{l ccc ccc ccc}
            \toprule
            \multirow{2}{*}{Method} & \multicolumn{3}{c}{N20A2} & \multicolumn{3}{c}{N50A3} & \multicolumn{3}{c}{N100A4} \\
            \cmidrule(lr){2-4} \cmidrule(lr){5-7} \cmidrule(lr){8-10}
            & HV$\uparrow$ & Gap & Time & HV$\uparrow$ & Gap & Time & HV$\uparrow$ & Gap & Time \\
            \midrule
            MOEA/D   & 0.201 & 16.71\% & 32.92m & 0.374 & 15.75\% & 2.58h  & 0.394 & 37.33\% & 4.56h \\
            MOGLS    & 0.214 & 11.57\% & 7.72h  & 0.207 & 53.38\% & 19.88h & 0.227 & 63.85\% & 38.43h \\
            NSGA-II  & 0.236 & 2.48\%  & 34.89m & 0.389 & 12.39\% & 2.47h  & 0.416 & 33.76\% & 4.28h \\
            NSGA-III & 0.235 & 3.06\%  & 45.37m & 0.386 & 13.18\% & 2.50h  & 0.402 & 35.94\% & 4.34h \\
            MO-PARCO    & 0.229 & 5.46\%  & 1.53m & 0.422 & 4.96\%  & 2.56m & 0.591 & 5.95\%  & 3.22m \\
            \midrule
            CAMO     & 0.240 & 0.91\%  & 10.55s & 0.439 & 1.02\%  & 20.77s & 0.624 & 0.65\%  & 55.31s \\
            CAMO-IA  & \textbf{0.242} & \textbf{0.00\%} & 16.96s & \textbf{0.444} & \textbf{0.00\%} & 59.52s & \textbf{0.628} & \textbf{0.00\%} & 5.44m \\
            \bottomrule
        \end{tabular}
    }
\end{table}

The experimental results summarized in Table \ref{tab:Comparison Results} show that CAMO consistently outperforms all baseline methods, with its advantage becoming more pronounced as the problem scale increases. Moreover, when equipped with the IA strategy, CAMO-IA achieves further performance improvements. While traditional MOEAs remain competitive in the small-scale N20A2 scenario, their performance degrades substantially on larger N50A3 and N100A4 instances. MO-PARCO remains competitive across all problem scales, yet consistently underperforms our CAMO. In terms of efficiency, neural solvers significantly reduce computation time.

\noindent\textbf{Generalization Analysis.}
To evaluate generalization, we test CAMO and all baselines on 100 randomly generated instances for each larger-scale configuration, including N100A6, N150A4, N150A6, N200A4, and N200A6, which are unseen during training. For fairness, MO-PARCO is evaluated using the checkpoint trained on problem size 100 to ensure its best performance. We set the HV reference points to [40, 21] for N100A6, [43, 32] for N150A4, [49, 29] for N150A6, [57, 41] for N200A4, and [62, 37] for N200A6. The results in Table~\ref{tab:Generalization} show that CAMO-IA achieves the highest HV in every case, demonstrating strong generalization to larger-scale MOMTSPs.

\begin{table*}[ht]
    \centering
    \caption{Generalization experiment results.}
    \label{tab:Generalization}
    
    \renewcommand{\arraystretch}{1.0}
    \setlength{\tabcolsep}{3pt}
    
    \resizebox{\textwidth}{!}{
        \begin{tabular}{l ccc ccc ccc ccc ccc}
        \toprule
        \multirow{2}{*}{Method} & \multicolumn{3}{c}{N100A6} & \multicolumn{3}{c}{N150A4} & \multicolumn{3}{c}{N150A6} & \multicolumn{3}{c}{N200A4} & \multicolumn{3}{c}{N200A6} \\
        \cmidrule(lr){2-4} \cmidrule(lr){5-7} \cmidrule(lr){8-10} \cmidrule(lr){11-13} \cmidrule(lr){14-16}
        & HV$\uparrow$ & Gap & Time & HV$\uparrow$ & Gap & Time & HV$\uparrow$ & Gap & Time & HV$\uparrow$ & Gap & Time & HV$\uparrow$ & Gap & Time \\
        \midrule
        MOEA/D   & 0.536 & 24.36\% & 4.51h & 0.348 & 50.41\% & 5.83h & 0.426 & 41.54\% & 5.97h & 0.265 & 64.02\% & 7.26h & 0.342 & 54.82\% & 7.47h \\
        MOGLS    & 0.393 & 44.49\% & 50.96h & 0.214 & 69.47\% & 52.39h & 0.289 & 60.36\% & 67.88h & 0.186 & 74.80\% & 65.65h & 0.243 & 67.86\% & 81.76h \\
        NSGA-II  & 0.528 & 25.42\% & 4.45h & 0.379 & 45.93\% & 5.91h & 0.416 & 42.94\% & 5.94h & 0.327 & 55.69\% & 7.53h & 0.348 & 53.97\% & 7.75h \\
        NSGA-III & 0.518 & 26.84\% & 4.51h & 0.356 & 49.17\% & 5.97h & 0.396 & 45.75\% & 6.29h & 0.288 & 61.04\% & 7.61h & 0.315 & 58.37\% & 7.82h \\
        MO-PARCO    & 0.669 & 5.47\%  & 3.46m & 0.664 & 5.22\%  & 4.95m & 0.699 & 4.05\%  & 5.49m & 0.682 & 7.55\%  & 6.35m & 0.720 & 4.71\%  & 7.23m \\
        \midrule
        CAMO     & 0.704 & 0.59\% & 1.39m & 0.698 & 0.53\% & 2.48m & 0.724 & 0.63\% & 3.39m & 0.734 & 0.48\% & 6.00m & 0.752 & 0.53\% & 7.15m \\
        CAMO-IA  & \textbf{0.708} & \textbf{0.00\%} & 7.62m & \textbf{0.701} & \textbf{0.00\%} & 16.44m & \textbf{0.729} & \textbf{0.00\%} & 23.16m & \textbf{0.738} & \textbf{0.00\%} & 35.42m & \textbf{0.756} & \textbf{0.00\%} & 49.06m \\
        \bottomrule
        \end{tabular}
    }
    \vspace{-4mm}
\end{table*}

\noindent\textbf{Benchmark  Evaluation.}
We further evaluate CAMO’s generalization on out-of-distribution benchmark instances adapted from TSPLIB~\cite{reinelt1991tsplib}. We denote these instances as “X$i$-A$j$”, where X is the TSPLIB instance name, $i$ is the number of nodes, and $j$ is the number of agents in our modified setting.
\begin{table}[ht]
\centering
\caption{Performance comparison between NSGA-II and CAMO on benchmark instances.}
\label{tab:benchmark_instance_results}

\renewcommand{\arraystretch}{1.0}
\setlength{\tabcolsep}{6pt}

\begin{tabular}{l cc cc}
\toprule
\multirow{2}{*}{Instance} & \multicolumn{2}{c}{Obj.Sum $\downarrow$} & \multicolumn{2}{c}{Obj.Max $\downarrow$} \\
\cmidrule(lr){2-3} \cmidrule(lr){4-5}
 & NSGA-II & CAMO & NSGA-II & CAMO \\
\midrule
ulysses22-A2  & 4.00 & \textbf{3.95} & \textbf{2.28} & 2.33 \\
swiss42-A2    & 5.61 & \textbf{5.14} & 3.11 & \textbf{2.84} \\
gr21-A3       & \textbf{3.78} & 3.78 & \textbf{2.01} & 2.08 \\
berlin52-A4   & 6.14 & \textbf{5.54} & 2.06 & \textbf{1.82} \\
att48-A5      & 5.84 & \textbf{5.48} & \textbf{1.91} & 1.93 \\
brazil58-A6   & 5.03 & \textbf{4.19} & 2.07 & \textbf{2.06} \\
eil76-A7      & 11.91 & \textbf{8.30} & 2.78 & \textbf{2.26} \\
pr76-A10      & 11.73 & \textbf{6.90} & 2.74 & \textbf{2.47} \\
kroA150-A15   & 28.76 & \textbf{9.49} & 3.44 & \textbf{2.27} \\
kroB200-A20   & 41.83 & \textbf{10.99} & 4.09 & \textbf{2.56} \\
\bottomrule
\end{tabular}
\vspace{-3mm}
\end{table}
For rigorous evaluation and fair comparison, we apply a unified preprocessing procedure to TSPLIB instances that normalizes all node coordinates into the $[0,1]^2$ unit square.
For instances that only provide edge-weight distance matrices (i.e., swiss42, gr21, and brazil58), we first reconstruct node coordinates using multidimensional scaling (MDS)~\cite{tenenbaum2000global}, and then apply the same normalization. We compare CAMO with NSGA-II in terms of Obj.Sum and Obj.Max, which denotes the minimum PF values along the two objectives, where smaller is better.

As shown in Table~\ref{tab:benchmark_instance_results}, CAMO achieves superior performance on most benchmark instances. In terms of Obj.Sum, CAMO obtains lower values than NSGA-II on 9 out of 10 instances, with significant gains on medium- and large-scale cases such as pr76-A10, kroA150-A15, and kroB200-A20. For Obj. Max, CAMO outperforms NSGA-II on 6 out of 10 instances, indicating that it also reduces makespan in most cases. These results demonstrate that CAMO maintains strong generalization on benchmark instances whose distributions differ from those used in training.

\begin{figure}[h]
    \captionsetup{skip=2pt}
    \centering
    \includegraphics[width=\linewidth]{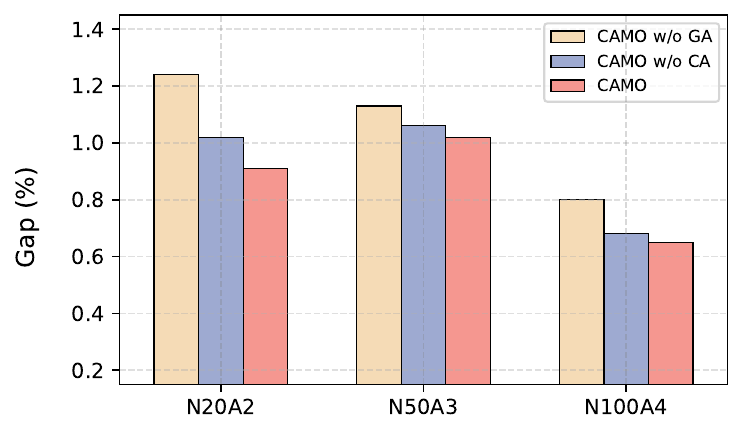}
    \caption{Ablation study results of CAMO and its variants. Lower values of Gap (\%) indicate better performance. }
    \label{fig:Ablation}
\end{figure}

\noindent\textbf{Ablation Study.} We conduct an ablation study to assess the contributions of key components in CAMO, including the GA module, the CA mechanism, and the customized sampling strategy. Note that we do not ablate the collaborative decoder, as removing it prevents CAMO from handling the multi-agent setting.
First, we replace the GA module with a residual connection and replace the conditional attention (CA) mechanism with standard self-attention, resulting in the variants CAMO w/o GA and CAMO w/o CA, respectively.
The performance of the original CAMO and its variants on N20A2, N50A3, and N100A4 is illustrated in Fig.~\ref{fig:Ablation}.
The results show that the performance of CAMO deteriorates significantly when any of these components is removed, confirming the effectiveness of each key component in the model architecture.
Next, we train a CAMO variant without the customized sampling strategy by fixing the training instances to the N100A4 configuration, denoted as CAMO (N100A4). 
We compare the training time and HV across problem scales (N50A3, N100A4, and N200A4). Table~\ref{tab:ablation_N100A4} shows that our customized sampling strategy reduces training time by more than 50\% while achieving competitive performance on all scales. It is only slightly inferior to CAMO (N100A4) on the N100A4 setting, demonstrating the effectiveness of our training strategy.

\begin{table}[ht]
\centering
\caption{CAMO vs. CAMO (N100A4).}
\label{tab:ablation_N100A4}

\renewcommand{\arraystretch}{1.0}
\setlength{\tabcolsep}{6pt}

\begin{tabular}{l c ccc}
\toprule
\multirow{2}{*}{Method} & \multirow{2}{*}{Training Time $\downarrow$} & \multicolumn{3}{c}{HV $\uparrow$} \\
\cmidrule(lr){3-5}
 &  & N50A3 & N100A4 & N200A4 \\
\midrule
CAMO (N100A4) & 168.62 h & 0.4308 & \textbf{0.6263} & 0.7339 \\
CAMO          & \textbf{70.42 h}  & \textbf{0.4394} & 0.6241 & \textbf{0.7340} \\
\bottomrule
\end{tabular}

\end{table}

\subsection{Real-World Robot Validation} We implemented our proposed CAMO framework on a physical robot
platform to validate its effectiveness in real-world robot-node planning scenarios. We deployed our algorithm on 3 mobile robots operating in an indoor environment where the robots were required to traverse 51 nodes, comprising one depot and 50 task nodes.
\begin{figure}[h]
    \centering
    
    \begin{subfigure}{0.48\columnwidth}
        \centering
        \includegraphics[width=\linewidth]{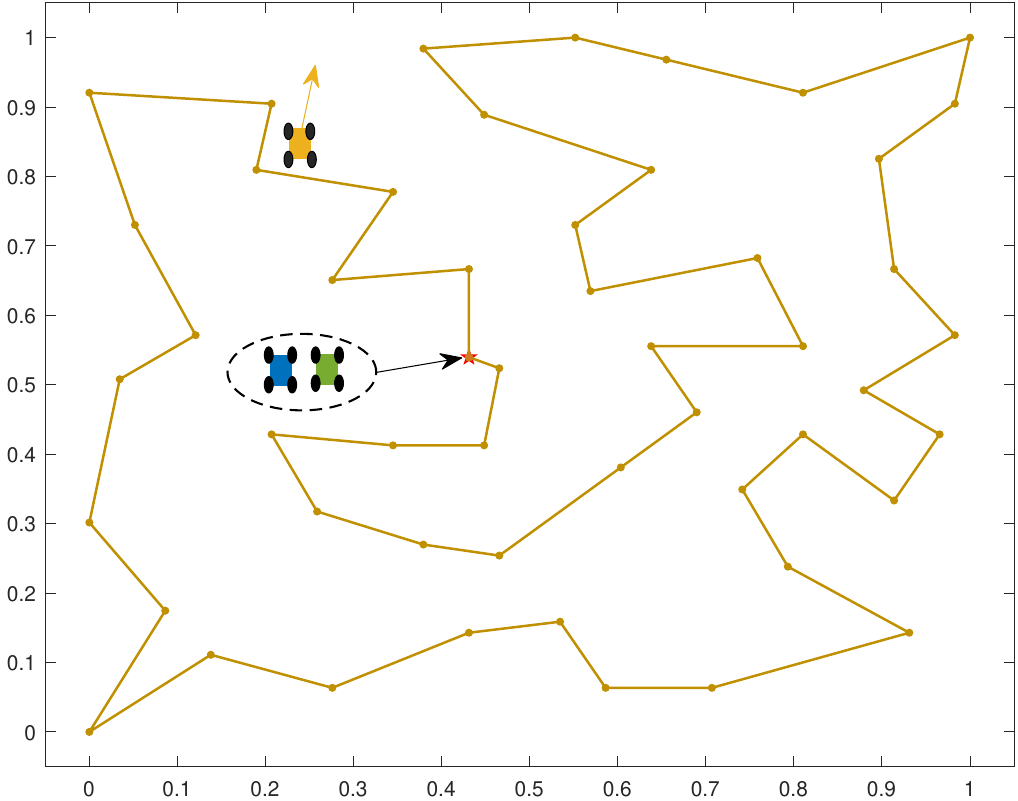}
        \caption{Simulation for $\lambda$=[1,0]}
        \label{fig:simu1}
    \end{subfigure}
    \hfill
    \begin{subfigure}{0.48\columnwidth}
        \centering
        \includegraphics[width=\linewidth]{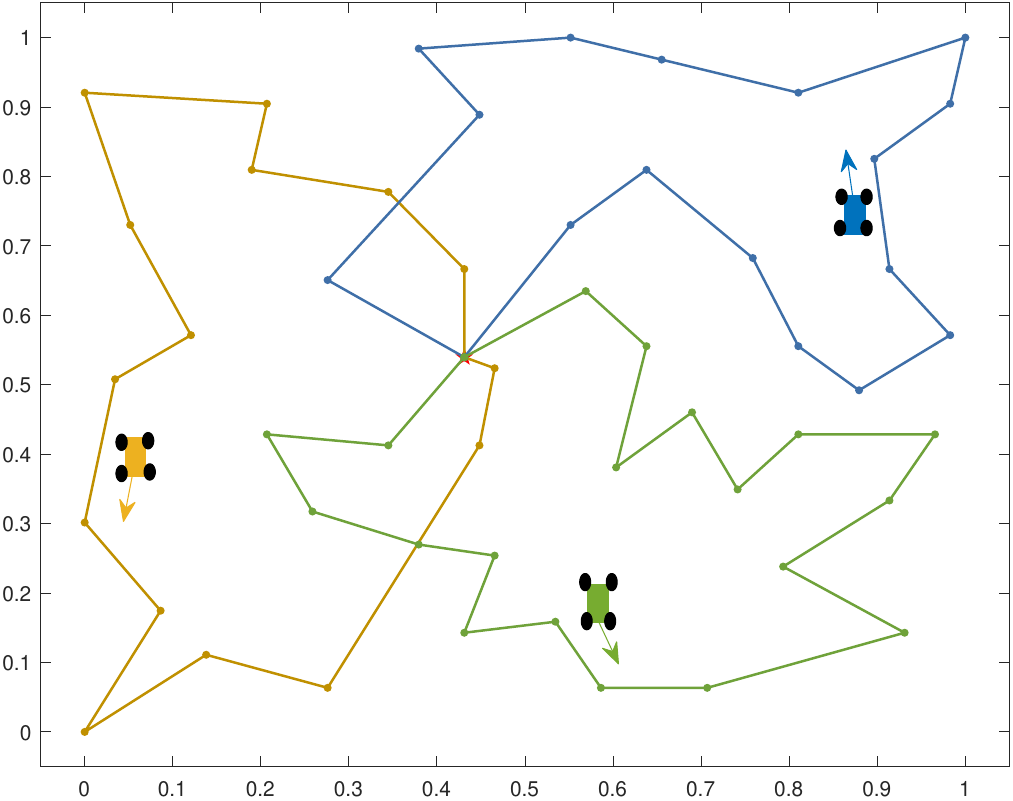}
        \caption{Simulation for $\lambda$=[0,1]}
        \label{fig:simu2}
    \end{subfigure}
    
    \vspace{0.5em}
    
    \begin{subfigure}{0.48\columnwidth}
        \centering
        \includegraphics[width=\linewidth]{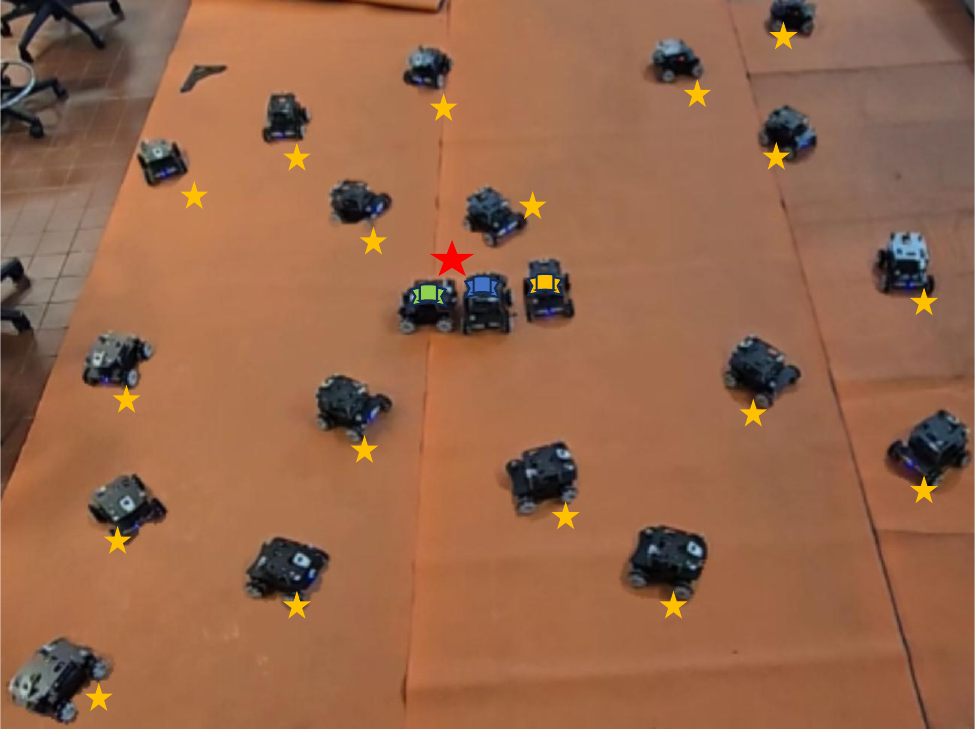}
        \caption{Real-world test for $\lambda$=[1,0]}
        \label{fig:case1}
    \end{subfigure}
    \hfill
    \begin{subfigure}{0.48\columnwidth}
        \centering
        \includegraphics[width=\linewidth]{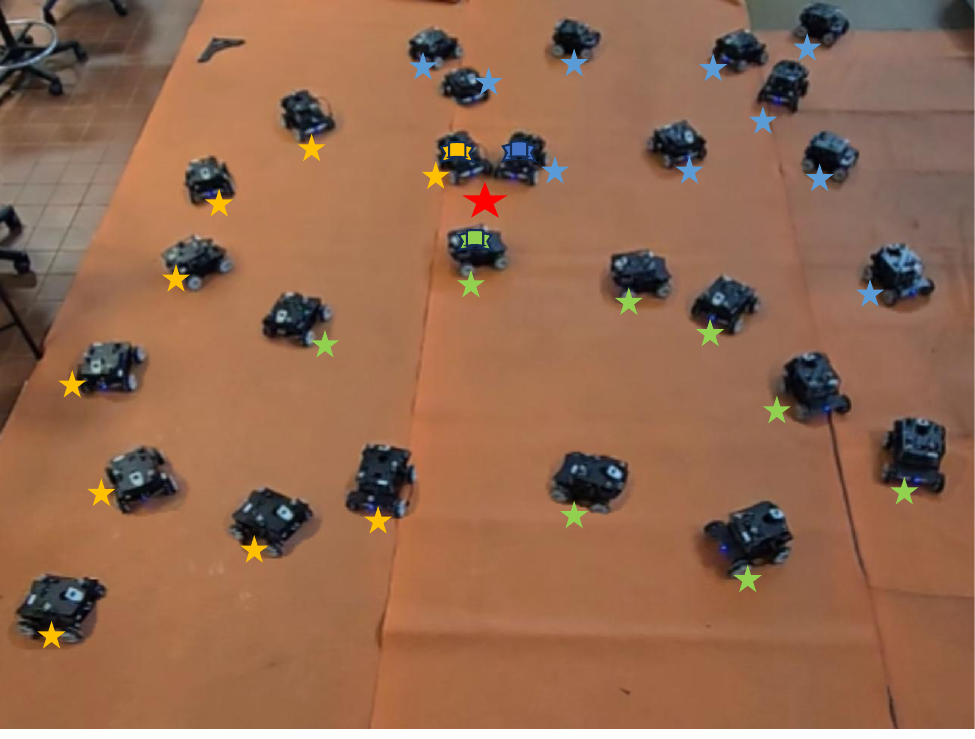}
        \caption{Real-world test for $\lambda$=[0,1]}
        \label{fig:case2}
    \end{subfigure}
    
    \caption{Experimental results in simulation (top) and real-world (bottom) scenarios.}
    \label{fig:exp}
\end{figure}
Fig.~\ref{fig:exp} shows our experimental simulation and real-robot results, where the robots successfully visited all 50 task nodes. The paths represent the solutions generated by our algorithm, corresponding to the fixed preference vectors $\lambda_1 = [1, 0]$ and $\lambda_2 = [0, 1]$, respectively, demonstrating CAMO's ability to effectively solve MOMTSP instances in practical settings.

\section{Conclusion}
In this paper, we propose CAMO, a conditional neural solver for MOMTSP. CAMO fuses preference and instance information via a conditional encoder and coordinates multi-agent decision making through a collaborative decoder. We further train CAMO with a REINFORCE-based objective over a mixed distribution of problem sizes to improve robustness to unseen agent and node configurations. Empirically, CAMO delivers strong performance, generalizes well to out-of-distribution instances and larger scales, and demonstrates practical applicability through real-world deployment on a mobile robot platform. In future work, we will investigate how to address the imbalanced objectives commonly encountered in multi-agent MOCOPs and explore solving large-scale MOMTSP instances through advanced generative models or divide-and-conquer principles.

\section*{ACKNOWLEDGMENT}


We used ChatGPT to assist in the linguistic refinement of the Introduction and Conclusion sections. 
The generated text was critically reviewed and revised by the authors to ensure alignment with the research findings and academic standards.

\bibliographystyle{IEEEtran}
\bibliography{ref}
\end{document}